# ReacLLaMA: Merging chemical and textual information in chemical reactivity AI models

Aline Hartgers[1], Ramil Nugmanov[2*], Kostiantyn Chernichenko[2] and Joerg Kurt Wegner[3]

[1]Department of Computer Science, KU Leuven, Celestijnenlaan 200A, 3001 Leuven, Belgium

[2]Janssen Research & Development, a division of Janssen Pharmaceutica N.V. Turnhoutseweg 30, Beerse, B-2340, Belgium

[3]Janssen Research & Development, LLC, 255 Main St, Cambridge, MA 02142, USA

*e-mail: rnugmano@its.jnj.com

## Abstract

Chemical reactivity models are developed to predict chemical reaction outcomes in the form of classification (success/failure) or regression (product yield) tasks. The vast majority of the reported models are trained solely on chemical information such as reactants, products, reagents, and solvents, but not on the details of a synthetic protocol. Herein incorporation of procedural text with the aim to augment the Graphormer reactivity model and improve its accuracy is presented. Two major approaches are used: training an adapter Graphormer model that is provided with a GPT-2-derived latent representation of the text procedure (ReacLLaMA-Adapter) and labeling an unlabeled part of a dataset with the LLaMA 2 model followed by training the Graphormer on an extended dataset (Zero-Shot Labeling ReacLLaMA). Both methodologies enhance the discernment of unpromising reactions, thereby providing more accurate models with improved specificity.

## Introduction

Accurate prediction of organic reaction yields is a complex task that is thought to revolutionize organic synthesis and is one of the areas where AI/ML techniques could accelerate drug development[1–3]. Chemists can conduct more promising chemical experiments in-vitro when data-driven methods can adequately predict yield in-silico, and thereby waste fewer valuable resources such as materials and time. These models can also aid in the automation of proposing promising reagents for synthesis: leaving this choice to a machine enables the incorporation of more knowledge than a human can ever acquire individually, and furthermore reduces bias on the chosen reagents. Reaction success and optimal condition predictions complement retrosynthetic software in a complete CASP pipeline, with a potential execution by robotic equipment resulting in fully automatic synthesis pipelines.[4]

Given the interesting use cases, building chemical reactivity models has been an often-tackled problem. This results in different ML approaches that include, among others, those utilizing SMILES[5] sequence analysis[3], graph neural networks (e.g., Graphormer[6,7], WLN[8]), introducing physical properties[9], and using molecular fingerprints[10–14]. Transformer models[15,16], which are specialized in sequence-to-sequence tasks, have proven themselves useful in processing reagents' to products' SMILES[17]. All these techniques involve only structured data to work with. Here we demonstrate augmenting the models with the procedural texts from Electronic Laboratory Notebooks (ELN) and implement it using two different approaches.

Organic synthesis ELNs represent an invaluable data source for training reactivity models due to, among other things, a significant number of experiments with negative or undesired outcomes which sharply contrast with the highly unbalanced public literature data that almost exclusively report the results of experiments with non-zero yields. Nevertheless, accurate modeling of yield in real-life chemical datasets is complicated by factors that are not captured in structured data, for example, equipment setup, isolation techniques, order of mixing chemicals, temperature, duration of the reaction, stirring speed, etc[18]. Therefore, in the present approach, the reaction outcome is modeled as a classification task where reactions with yields less than 5% are considered as having a negative outcome. Although using 5% as a cutoff value is arbitrary, it is dictated by practical considerations in a medicinal chemistry setting. Another aspect of real-life data is incompleteness: about 30% of the reactions in the dataset we used do not contain a reaction outcome in form of structured data. This, however, provides an opportunity for non-structured text processing (see below), as indications of the reaction outcome are typically present in procedural texts.

In general, we aimed to build a model that incorporates procedural text during training, but the inference (and hence testing) should be done using the structured chemical information only. Such a framework is motivated by two practical aspects: firstly, a model useful in a wet chemistry lab should not require a procedure for predicting the outcome of the reaction, and, secondly, such setup allows avoiding target leaks or model adversarial attacks[19] from a procedural text. To merge chemical and textual information, we propose two approaches. First, we present a model inspired by the *LLaMA-Adapter V2*[20] that can be utilized in multi-modal contexts. In our implementation, *GPT-2*[21] (raw text) and Graphormer (reaction structure) are combined to predict yield. This results in *ReacLLaMA-Adapter* (reaction + LLaMA-Adapter). Our second approach, *Zero-Shot Labeling ReacLLaMA* (ZSL ReacLLaMA), is a method to perform automatic label extraction from procedural texts based on zero-shot learning to increase the size of a labeled dataset for yield prediction models. For this, the language model *LLaMA 2*[22] retrieves embeddings from the last token of every text procedure, and we subsequently use these to train a neural network that predicts reaction success labels. Because of automatically assigned labels, the size of the labeled dataset available for training is increased by 43% which improves the performance of a core Graphormer model in the yield predicting task.

## Methods

### Dataset

| | |
|---|---|
| **Total** | 2 446 167 |
| **Near-empty procedures** | 5456 (0.2%) |
| **Labeled data** | 1 713 237 (70%) |
| **Unlabeled data** | 732 930 (30%) |
| **Positive labels (out of labeled)** | 1 270 982 (75%) |
| **Negative labels (out of labeled)** | 442 255 (25%) |

*Table 1: ELN data statistics*

All experiments are conducted on internal proprietary data, a part of J&J medicinal chemistry ELNs. The experiments containing more than one product were split, ultimately producing a dataset containing around 2.4 million single-product reactions. Structural data is processed as described before[23,24]. Reactants and products were separated from reagents, solvents, and catalysts (RSCs) and are treated by the model as graphs (see below), whereas RSCs are used as categorical variables. In this context, reactants are starting compounds whose atoms constitute the product whereas this is not the case for RSCs. In the dataset, we found 1625 unique RSCs. 30% of the reactions are unlabeled, i.e., they do not contain information about yield or reaction outcome (failure or success). A successful (positively labeled) experiment contains a yield of the product equal to or above 5%, whereas a failed reaction has a lower yield or a negative reaction outcome label. Within the labeled experiments, the ratio of positive to negative labels is approximately 3:1. 5456 of the procedural texts are considered near-empty (less than five words long). It should however be stated that some of them still contain useful information, for example when the procedure states 'failed'. The non-structured texts are complimented with the information from the structured fields using '##' tags (See Table 2).

| Tag | Example sample | Comments |
| --- | --- | --- |
| ##technology## | Library | Single, high-throughput experiment, a parallel library, or a technology (flow, etc.) |
| ##procedure## | A, B, D, F in G were stirred … | Compounds in most cases labeled alphabetically |
| ##comments## |  | Optional comments given by chemists in free text format |
| ##product## | P1 | Label of the product in procedure text |
| ##yield## | 4.0% | Can be empty |
| ##label## | neg | "neg", "pos", or empty |

Table 2: Procedure text schema (prepared for training)

Within this dataset, a subset of Pd-catalyzed reactions is also considered separately. This subset includes all reactions where Pd appears as a catalyst in whichever form (complex, enzyme, etc.). Pd-catalyzed reactions, such as Buchwald-Hartwig amination and Suzuki-Miyaura coupling, are frequently used in organic synthesis in general and medicinal chemistry and therefore are of particular interest. This subset contains 30108 single-product reactions.

## The core (baseline) Graphormer model

The core model for both text-enhancing approaches is based on a Graphormer architecture that processes reactants and products as molecular graphs. In previous work[6], we introduced a BERT-based reaction encoder neural network, which has two parts: the first part is a direct molecular graph encoder network inspired by a Graphormer architecture[7] used as an atoms-in-molecule embedding generator and the second part is a BERT[15] network for reactants-to-products relationships learning. We demonstrated a state-of-the-art performance of this architecture in the atom-to-atom mapping task, one of the important tasks in reactivity modeling.

Since then, we have found that the Graphormer part of the network with enabled attention between reactants and product molecules is on its own sufficient for many tasks. Similar to the previous architecture, the new model encodes atoms and total neighbors count, which include implicit hydrogens (previously called the centrality embedding), as learnable embedding vectors (Figure 1). The role encoder from the previous architecture (previously called the sentence embedding) is moved to the beginning of the network and is modified to only bias product atom embeddings. This bias is found sufficient to separate reactants and products. The distance encoder (previously called the spatial encoder) that codes pairwise distances between atoms and is used as a bias to atoms attention matrices in the multi-head attention block of the transformer is kept the same. However, in the new architecture cross-molecular and cross-component (ions of salts, etc.) biases are disabled.

RSCs of reactions in the new architecture are coded as a sequence of categorical identifiers. Attention from molecules to RSCs is disabled by attention masking (see Figure 1). Thus, we keep only reagents to structure unbiased attention. A similar masking approach is described in T5 as a mask with prefix[25].

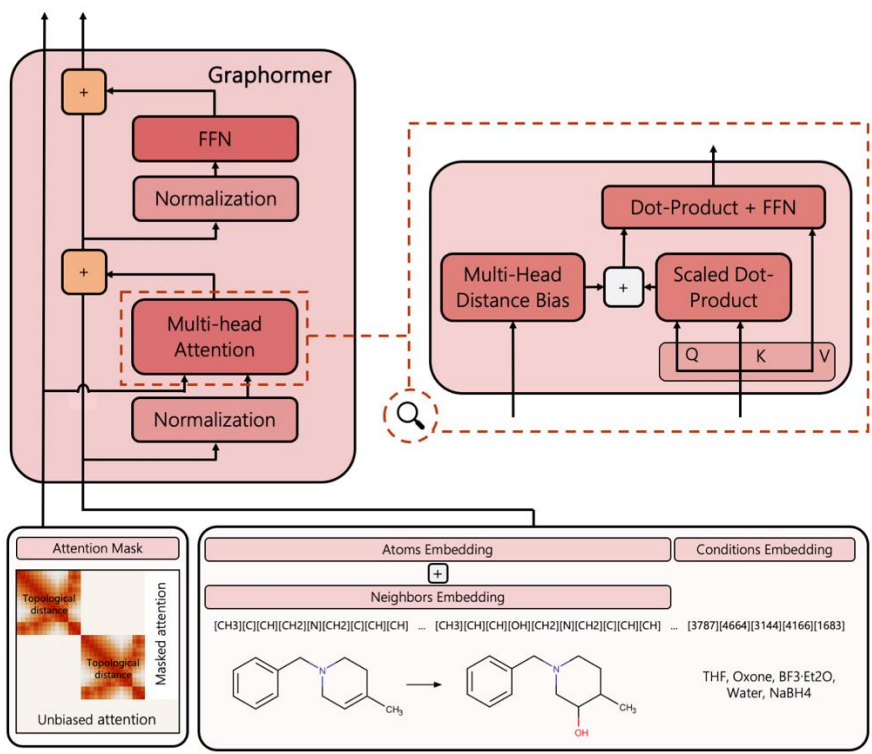

*Figure 1: Baseline Graphormer model architecture and reaction input*

The Graphormer is pretrained on a Masked Language Modeling (MLM) task to restore randomly hidden atom types in molecules of reactants and products and reagent categorical labels. The core idea behind MLM involves seeking out matching atoms within both the reactants and the products as described before[6,17]. The core model that trained solely on structured chemical (reactants, products, RSCs) data was used as a baseline model. For all experiments, the J&J ELN data points are partitioned temporally, with data collected in the year 2022 being allocated as the test set (~10%), while all data recorded prior to that year is designated as the training set. This partitioning strategy is implemented to facilitate predictive modeling that produces representative predictions.

## ReacLLaMA-Adapter

In ReacLLaMA-Adapter, GPT-2 is added to improve model prediction by incorporating textual information in the following way. The Graphormer is pretrained by an MLM task as described above and GPT-2 is pretrained on chemical procedural texts to adjust to the specific linguistic domain. It should be noted that this is a second pretraining for GPT-2, as it was already trained beforehand on general text corpora. Next, the technique of adaptation is applied as described by Gao et al.[20] where the two modalities, a reaction and a text, are processed together in adaptation layers (see Figure 2b). In such a layer, the reaction goes through a regular Transformer layer with unfrozen weights which consists of multi-head attention, normalization and a feedforward network (FFN). The output of this reaction sublayer is then upsampled to fit GPT-2 embedding dimensionality and added as an input to the adapted multi-head attention of the textual sublayer. All other blocks in this sublayer are standard Transformer blocks with frozen weights.

This adapted multi-head attention block (see Figure 2c) differs from regular multi-head attention blocks. Here, instead of the usual single scaled dot-product attention, there is one for each modality (text and reaction). The query originates from the text, and for both modalities this query is compared to the available information in the individual sources with their respective frozen keys and values through scaled dot-product attention. The attention mask allows all product atoms to see all reactant atoms, whereas the RSCs can both look at all reactant atoms and all other RSCs. If there is useful information in the reaction part, the trainable zero-initialized gate opens, and it is added to the result.

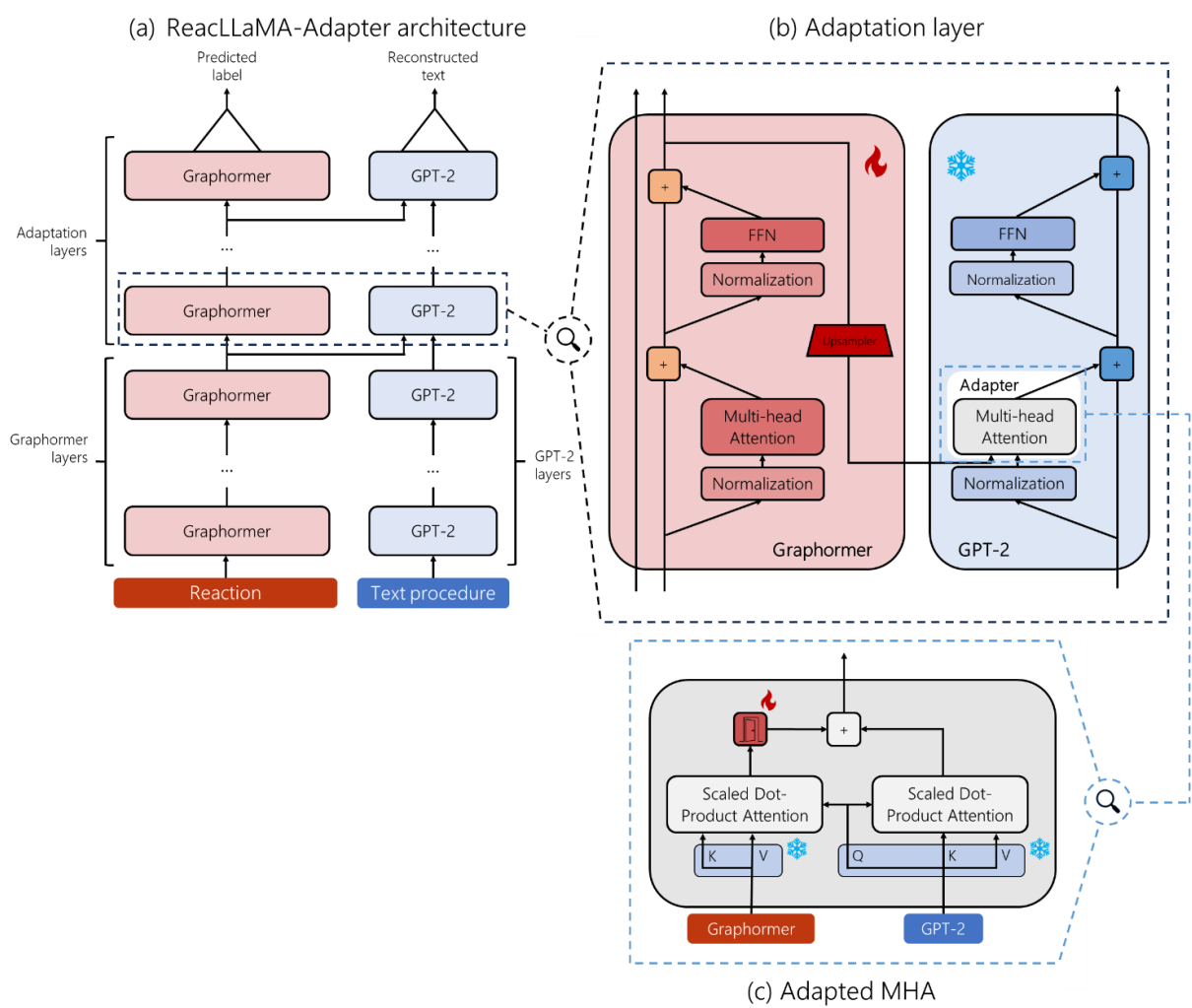

Figure 2: (a) Full ReacLLaMA-Adapter architecture with separate Graphormer layers, GPT-2 layers and adaptation layers where both modalities are combined. In this architecture, the choice of the number of each kind of layer can be varied. (b) Schema of a ReacLLaMA-Adapter adaptation layer where both a reaction and textual procedure are processed with the technique of adaptation. A Transformer architecture forms the base of both the reaction and textual sublayer, with multi-head attention, normalization and a feedforward network (FFN). The color indicates whether weights are frozen (blue) or unfrozen (red) in that

*part of the network. (c) Schema of an adapted multi-head attention (MHA) block. The query from the textual part is used for the scaled dot-product attention for both modalities. If the scaled-dot product of the chemical reaction proves useful during optimization, the zero-initialized gate opens. Here again, color indicates whether weights are frozen (blue) or unfrozen (red).*

The full ReacLLaMA-Adapter architecture (see Figure 2a) consists of Graphormer layers, GPT-2 layers and adaptation layers. The amount of each kind of layer can be chosen freely, however with restrictions when using a pretrained *Graphormer* or Large Language Model. For example, here the combined number of layers in the GPT-2 layers and adaptation layers should equate to the total number of layers present in the original GPT-2 architecture. Our choice of GPT-2 as the language model is based on the available infrastructure as well as on the nature of the task at hand. Even with frozen weights for the language model, it was not feasible to use a larger model and still have the flexibility to do multiple design iterations.

For *ReacLLaMA-Adapter* training, we used the following setup: 6 NVIDIA V100 32GB GPUs, a distributed data-parallel multi-GPU training strategy, an effective batch size 180, a learning rate $8 \times 10^{-4}$, GPT-2 (124M parameters), 8 Graphormer layers with 16 heads and an embedding size 256. All Graphormer layers used for adaptation. For yield prediction binary cross-entropy loss and for text prediction cross-entropy loss without any loss regularization are used, respectively. This training takes 15 hours, whereas a baseline Graphormer model takes 3.5 hours to train.

### Zero-Shot Labeling ReacLLaMA

*LLaMA 2*[22] (13B) is employed in the first step of the labeling process to perform zero-shot learning. This state-of-the-art model with a Transformer architecture is released by Meta AI and is already trained on more than 2 trillion tokens of general and scientific texts. For labeling, the embeddings for the last token of all ELN procedures are extracted. A single hidden layer neural network, with GELU[26] as an activation function and followed by a sigmoid layer, is subsequently trained to do proper labeling with binary cross-entropy loss. This trained model is used to label the unlabeled data points. As data for this experiment, a random 80:20 train-to-test ratio split is made within the labeled dataset. After 24h, LLaMA 2 extracted the embeddings of all ELN procedures. The shallow neural network used for labeling was trained on 1 NVIDIA A100 40GB GPU (batch size 3000, learning rate $1 \times 10^{-4}$). This trained model is then directly applied to the unlabeled data, as both training loss and validation loss values on the test set are comparable which implies no extra training methods are needed. The calculated labels are assigned to the unlabeled data points following three different strategies. The labels are (1) given as a probability of being positive, (2) given as a probability of being positive, however with cutoff thresholds outside of which labels are not supplied, and (3) as a binarized version of the predicted probabilities (see Figure 3).

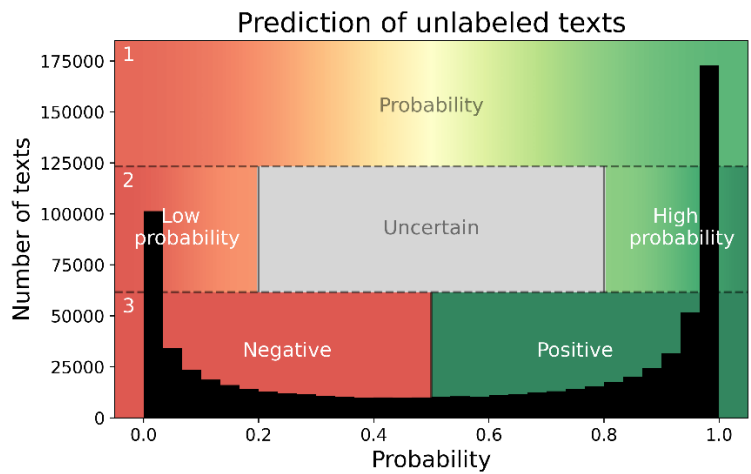

*Figure 3: A histogram of the predicted labels of textual procedures as probabilities, where 0 represents the highest probability of being negative and 1 the highest probability of being positive. The three labeling possibilities are visualized by the three horizontal bands, where (1) represents labeling as continuous probabilities, (2) represents labeling as continuous probabilities if a level of certainty is reached (here, outside of 20%-80%), and (3) as binarized versions of the labels.*

Finally, the entire dataset extended with the records previously unlabeled and now labeled by ZSL-ReacLLaMA serves for training a core Graphormer model on reaction success prediction with binary cross-entropy loss (a batch size 180, a learning rate $1 \times 10^{-4}$). Three models different by the form of assigned labels (Figure 3) are obtained and results of their performance serve to assess the labeling method. For the second labeling method, the chosen cut-off thresholds for the labels are 20%-80% and 5%-95%. It should be noted that such results are obtained with unoptimized Graphormer's hyperparameters as their optimization is out of the scope of the present work.

# Results

## ReacLLaMA-Adapter

Both balanced accuracy and specificity improve, by 0.45% and 1.51% respectively for the adapter model relatively to the baseline Graphormer (see Table 4). There is however a minimal decrease in sensitivity (0.6%). The difference in the performance of Pd-catalyzed reactions is rather similar: there is an improvement of balanced accuracy and specificity, respectively by 0.34% and 1.28%, while there is a decrease in sensitivity (0.59%).

We attribute the increase in the specificity of ReacLLaMA-Adapter to the pretraining of GPT-2 on all the procedures from the dataset, both labeled and unlabeled experiments. Presumably, this increases the knowledge about negative examples, because, as the ZSL ReacLLaMA labeling study reveals (vide infra), the proportion of negative examples is higher in unlabeled than in labeled reactions, and thus those negative examples could influence ReacLLaMA-Adapter. For example, when negative text is inputted while no label is provided. Here, only text loss is active and since GPT-2 weights are frozen, all information seeps through to the Graphormer. The incorporation of a more modern large language model might strengthen this ability even further.

## Zero-Shot Labeling ReacLLaMA

When performing a strict binary classification between the two possible classes, 58% of the automatically assigned labels are positive and 42% negative. The combination of the two datasets ("Combined labels") reduces the class imbalance present in ELN with only original labels, as can be seen in Table 3.

|  | Positive | Negative |
| --- | --- | --- |
| (1) Original labels | 75% | 25% |
| (2) ZSL ReacLLaMA labels | 58% | 42% |
| (3) Combined labels | 69% | 31% |

*Table 3: Distribution of binary labels in datasets.*

Generally, the network is rather certain when it comes to predicting class labels, as labels tend to be close to either 0 (negative) or 1 (positive) (see Figure 3). This can be attributed to the nature of the binary cross-entropy loss function. When measuring performance on the test set, the generated labels show 98% sensitivity and 85% specificity, yielding 91% balanced accuracy.

Some correlations are observed between the labels and the lengths of procedural texts. Figure 4 shows the testing performance of labeling for procedures of different lengths. Average labels are significantly lower for shorter texts growing steadily with the length until reaching a plateau. It is remarkable that whereas sensitivity does not show a clear trend, specificity and hence balanced accuracy significantly decline with increasing text length. Such observations are understandable: a failed reaction is often detected at an early stage prior to a product isolation attempt that results in a short procedure. For a successful reaction, a procedure may involve a description of purification steps or product characterization that increases its length. Very lengthy procedures may describe the multistage product isolation operations which may result in an overall negative outcome (<5% yield) since every stage is associated with losses. Eventually, the dropping sensitivity for the long procedures is not surprising considering that this sentiment analysis task is performed on purely textual information without the involvement of structural chemical inputs[27]. Besides, LLaMA 2 may not be able to capture field-specific nuances of medicinal chemical texts without additional pretraining.

The average label and the predicted average label in each text length class are compared to identify possible model assumptions. As the model is inclined to produce a similar average label for different length classes, it can be concluded that the label is not (at least solely) based on text length. Analysis of the procedural text lengths showed that unlabeled texts are on average shorter than the labelled ones (Figure 5). Although there is no reason to assume homogeneity in the nature of the procedures and their corresponding performance between the test set and the set of unlabeled data, we presume that the performance of labeling the unlabeled set is at least equally good as that of the test set, because texts in the unlabeled data set are generally shorter.

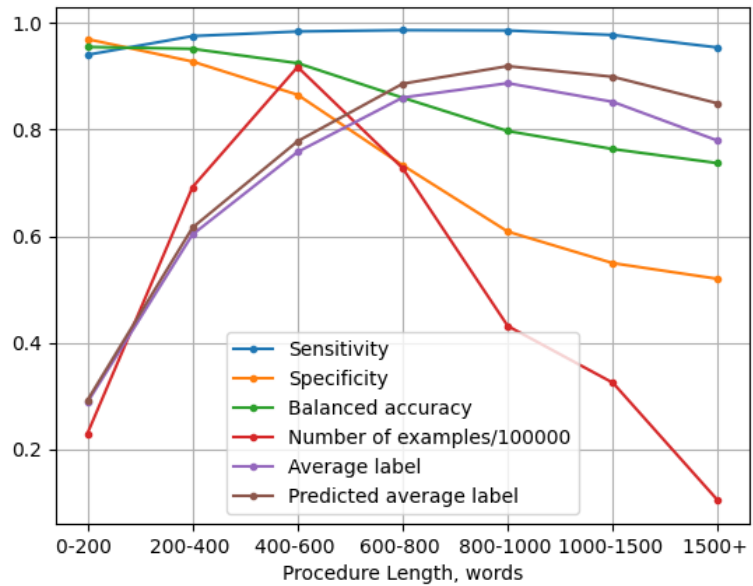

Figure 4: Performance metrics for textual procedures of different lengths when applying ZSL labeling on the validation dataset. The average label is the mean of all correct binary labels in the validation dataset, the predicted average label is the mean of the predicted binary labels.

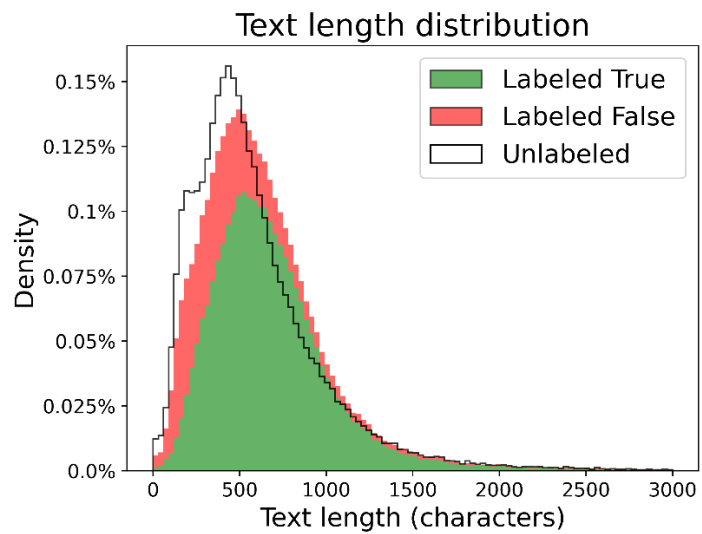

Figure 5: Distribution of textual procedures lengths (in characters) with cutoff at 3000

Below we investigate the influence of the new labels on the yield prediction model, the Graphormer. The chosen cut-off thresholds for the labels, 20%-80%, and 5%-95%, appoint labels to respectively 72% and 44% of previously unlabeled data. The cut-off means removing data points with predicted labels in threshold intervals. Procedures without labels are not used for training the model.

|  |  | Sensitivity | Specificity | Balanced accuracy | ROC AUC |
|---|---|---|---|---|---|
| **Baseline Graphormer** | | 93.27 | 54.65 | 73.96 | 85.53 |
| **ReacLLaMA-Adapter** | | 92.67 | 56.16 | 74.41 | 85.50 |
| **ZSL ReacLLaMA** | Binary | 90.94 | 56.39 | 73.67 | 84.26 |
| | Threshold 0.2-0.8 (60% confidence) | 91.35 | 57.96 | 74.66 | 85.10 |
| | Threshold 0.05-0.95 (90% confidence) | 92.30 | 56.46 | 74.38 | 85.31 |
| | Probability weighted | 90.56 | 59.06 | 74.81 | 84.89 |

*Table 4: Reactivity models performance metrics with confidence calculated as described in our previous work[3] (overall dataset)*

The results of yield prediction on the validation data set can be seen in Table 4. When compared to the baseline, specificity improved by 4.41% in the best case, which is for labeling with probabilities of being positive. However, sensitivity drops 2.71%. We observe this tendency as a gradient going from discrete (binary) to continuous (probability) labeling. The reduction in ELN class imbalance is a probable cause for this phenomenon. When specifically analyzing results for Pd-catalyzed reactions (see Table 5), trends are similar but more pronounced: there is a rise in specificity (5.15%) and balanced accuracy, and a drop in sensitivity (4.34%).

When comparing probability-weighted labels with binary labels, the numbers strongly indicate that binarizing the labels is detrimental to model performance. It gives rise to the assumption that uncertainty preserved in probabilities as weights better conserves information in the model. The results for the thresholds reinforce this suspicion, where labels that contain a certain amount of doubt are removed.

With the objective of yield prediction in mind, continuous labeling with probabilities as class weights performs best. Not identifying positive examples is unfortunate but bears less consequence than not identifying negatives since conducting unpromising experiments is a waste of resources.

|  |  | Sensitivity | Specificity | Balanced accuracy | ROC AUC |
|---|---|---|---|---|---|
| **Baseline Graphormer** | | 91.82 | 59.92 | 75.87 | 86.02 |
| **ReacLLaMA-Adapter** | | 91.23 | 61.20 | 76.21 | 85.83 |
| **ZSL ReacLLaMA** | Binary | 87.39 | 63.32 | 75.36 | 84.10 |
| | Threshold 0.2-0.8 (60% confidence) | 88.41 | 63.86 | 76.14 | 85.07 |
| | Threshold 0.05-0.95 (90% confidence) | 90.15 | 62.24 | 76.19 | 85.50 |
| | Probability weighted | 87.48 | 65.07 | 76.28 | 84.84 |

*Table 5: Reactivity models performance metrics for palladium-catalyzed reactions with confidence calculated as described in our previous work[3].*

To further substantiate the positive impact of labeled data extension, we conducted an analysis involving various thresholds for predicted probability binarization for the baseline model and the best ZSL ReacLLaMA with combined labels. In Figure 6, we observe that balanced accuracy maintains a consistent margin across a broad spectrum of thresholds. Consequently, we can infer that this sustained improvement is not attributed to a biasing of predictions towards the negative outcomes.

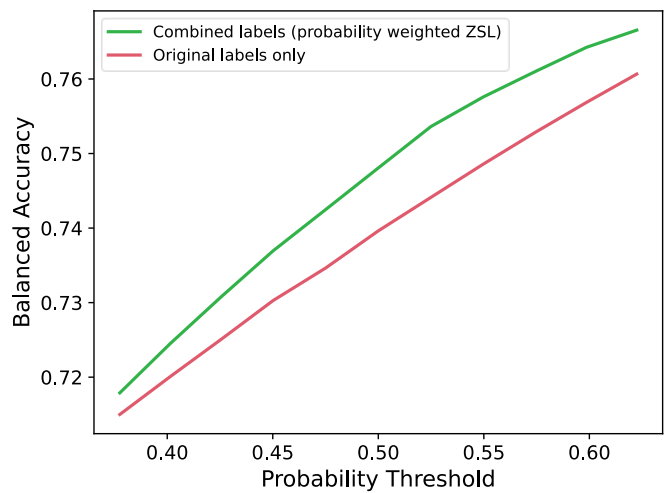

*Figure 6: Relationship between the positive label probability threshold and balanced accuracy for original labels and combined labels (probability weighted ZSL)*

## Conclusion

In this work we investigate two methods to improve chemical reactivity models for yield (reaction success) prediction with procedural texts as potential sources of additional information. The first proposal introduces ReacLLaMA-Adapter, an adapter Graphormer model accompanied by a GPT-2-derived latent representation of the text procedure, and the second is Zero-Shot Labeling ReacLLaMA (ZSL ReacLLaMA) that involves a zero-shot labeling method for those ELN records where reaction outcome is not captured as structured data.

For both cases, we designed our studies in such a way as to prevent potential target leakage from procedural texts, thus the inference and testing were performed only with the structured chemical data as input. As a result, the changes in performance in comparison to the baseline model were rather limited, however, we observed a consistent increase in specificity, a decrease in sensitivity, and moderate growth of balanced accuracy. The same trends were observed for the subset of Pd-catalyzed reactions.

LLaMA 2, with weights pretrained on general text corpora, is deemed sufficient to capture the result of the ELN procedure as a zero-shot learning task. The generated labels show a remarkable balanced test accuracy of 91%. These labels are incorporated in ZSL ReacLLaMA in three different manners: as continuous numbers (probabilities), as probabilities within a degree of confidence, and as binary labels. Since identifying both promising and unpromising reactions is equally important, continuous labels outperform all other versions of labeling, as seen by increased balanced accuracy and specificity while not decreasing sensitivity to a significant extent. The exposure to additional, previously unlabeled data and reduction in ELN class unbalance is a probable cause of this shift.

ReacLLaMA-Adapter shows the same tendencies in performance as ZSL ReacLLaMA, however to a lesser extent. Here, improvement might be also associated with exposure of the model to a larger dataset, because GPT-2 model was pretrained on procedures from both labeled and unlabeled reactions and later such information on negative examples might seep through to the Graphormer through the adapted multi-head attention block.

## References

(1) Struble, T. J.; Alvarez, J. C.; Brown, S. P.; Chytil, M.; Cisar, J.; DesJarlais, R. L.; Engkvist, O.; Frank, S. A.; Greve, D. R.; Griffin, D. J.; Hou, X.; Johannes, J. W.; Kreatsoulas, C.; Lahue, B.; Mathea, M.; Mogk, G.; Nicolaou, C. A.; Palmer, A. D.; Price, D. J.; Robinson, R. I.; Salentin, S.; Xing, L.; Jaakkola, T.; Green, William. H.; Barzilay, R.; Coley, C. W.; Jensen, K. F. Current and Future Roles of Artificial Intelligence in Medicinal Chemistry Synthesis. *J Med Chem* **2020**, *63* (16), 8667–8682. https://doi.org/10.1021/acs.jmedchem.9b02120.

(2) Madzhidov, T. I.; Rakhimbekova, A.; Afonina, V. A.; Gimadiev, T. R.; Mukhametgaleev, R. N.; Nugmanov, R. I.; Baskin, I. I.; Varnek, A. Machine Learning Modelling of Chemical Reaction Characteristics: Yesterday, Today, Tomorrow. *Mendeleev Communications* **2021**, *31* (6), 769–780. https://doi.org/10.1016/J.MENCOM.2021.11.003.

(3) Neves, P.; McClure, K.; Verhoeven, J.; Dyubankova, N.; Nugmanov, R.; Gedich, A.; Menon, S.; Shi, Z.; Wegner, J. K. Global Reactivity Models Are Impactful in Industrial Synthesis Applications. *J Cheminform* **2023**, *15* (1), 20. https://doi.org/10.1186/s13321-023-00685-0.

(4) Coley, C. W.; Thomas, D. A.; Lummiss, J. A. M.; Jaworski, J. N.; Breen, C. P.; Schultz, V.; Hart, T.; Fishman, J. S.; Rogers, L.; Gao, H.; Hicklin, R. W.; Plehiers, P. P.; Byington, J.; Piotti, J. S.; Green, W. H.; John Hart, A.; Jamison, T. F.; Jensen, K. F. A Robotic Platform for Flow Synthesis of Organic Compounds Informed by AI Planning. *Science (1979)* **2019**, *365* (6453). https://doi.org/10.1126/science.aax1566.

(5) Weininger, D. SMILES, a Chemical Language and Information System. 1. Introduction to Methodology and Encoding Rules. *J Chem Inf Comput Sci* **1988**, *28* (1), 31–36. https://doi.org/10.1021/ci00057a005.

(6) Nugmanov, R.; Dyubankova, N.; Gedich, A.; Wegner, J. K. Bidirectional Graphormer for Reactivity Understanding: Neural Network Trained to Reaction Atom-to-Atom Mapping Task. *J Chem Inf Model* **2022**, *62* (14), 3307–3315. https://doi.org/10.1021/acs.jcim.2c00344.

(7) Ying, C.; Cai, T.; Luo, S.; Zheng, S.; Ke, G.; He, D.; Shen, Y.; Liu, T.-Y. Do Transformers Really Perform Badly for Graph Representation? In *Advances in Neural Information Processing Systems*; Beygelzimer, A., Dauphin, Y., Liang, P., Vaughan, J. W., Eds.; 2021.

(8) Coley, C. W.; Jin, W.; Rogers, L.; Jamison, T. F.; Jaakkola, T. S.; Green, W. H.; Barzilay, R.; Jensen, K. F. A Graph-Convolutional Neural Network Model for the Prediction of Chemical Reactivity. *Chem Sci* **2019**, *10* (2). https://doi.org/10.1039/c8sc04228d.

(9) Davies, J. C.; Pattison, D.; Hirst, J. D. Machine Learning for Yield Prediction for Chemical Reactions Using in Situ Sensors. *J Mol Graph Model* **2023**, *118*, 108356. https://doi.org/10.1016/J.JMGM.2022.108356.

(10) Probst, D.; Schwaller, P.; Reymond, J.-L. Reaction Classification and Yield Prediction Using the Differential Reaction Fingerprint DRFP. *Digital Discovery* **2022**, *1* (2), 91–97. https://doi.org/10.1039/D1DD00006C.

(11) Afonina, V. A.; Mazitov, D. A.; Nurmukhametova, A.; Shevelev, M. D.; Khasanova, D. A.; Nugmanov, R. I.; Burilov, V. A.; Madzhidov, T. I.; Varnek, A. Prediction of Optimal Conditions of Hydrogenation Reaction Using the Likelihood Ranking Approach. *Int J Mol Sci* **2021**, *23* (1), 248. https://doi.org/10.3390/ijms23010248.

(12) Nugmanov, R. I.; Madzhidov, T. I.; Khaliullina, G. R.; Baskin, I. I.; Antipin, I. S.; Varnek, A. A. Development of "Structure-Property" Models in Nucleophilic Substitution Reactions Involving Azides. *Journal of Structural Chemistry* **2014**, *55* (6), 1026–1032. https://doi.org/10.1134/S0022476614060043.

(13) Madzhidov, T. I.; Gimadiev, T. R.; Malakhova, D. A.; Nugmanov, R. I.; Baskin, I. I.; Antipin, I. S.; Varnek, A. A. Structure–Reactivity Relationship in Diels–Alder Reactions Obtained Using the Condensed Reaction Graph Approach. *Journal of Structural Chemistry* **2017**, *58* (4), 650–656. https://doi.org/10.1134/S0022476617040023.

(14) Polishchuk, P.; Madzhidov, T.; Gimadiev, T.; Bodrov, A.; Nugmanov, R.; Varnek, A. Structure–Reactivity Modeling Using Mixture-Based Representation of Chemical Reactions. *J Comput Aided Mol Des* **2017**, *31* (9), 829–839. https://doi.org/10.1007/s10822-017-0044-3.

(15) Devlin, J.; Chang, M. W.; Lee, K.; Toutanova, K. BERT: Pre-Training of Deep Bidirectional Transformers for Language Understanding. *NAACL HLT 2019 - 2019 Conference of the North American Chapter of the Association for Computational Linguistics: Human Language Technologies - Proceedings of the Conference* **2018**, *1*, 4171–4186.

(16) Vaswani, A.; Shazeer, N.; Parmar, N.; Uszkoreit, J.; Jones, L.; Gomez, A. N.; Kaiser, Ł.; Polosukhin, I. Attention Is All You Need. *Adv Neural Inf Process Syst* **2017**, *2017-December*, 5999–6009. https://doi.org/10.48550/arxiv.1706.03762.

(17) Schwaller, P.; Hoover, B.; Reymond, J.-L.; Strobelt, H.; Laino, T. Extraction of Organic Chemistry Grammar from Unsupervised Learning of Chemical Reactions. *Sci Adv* **2021**, *7* (15), eabe4166. https://doi.org/10.1126/sciadv.abe4166.

(18) Voinarovska, V.; Kabeshov, M.; Dudenko, D.; Genheden, S.; Tetko, I. When Yield Prediction Does Not Yield Prediction: An Overview of the Current Challenges. **2023**. https://doi.org/10.26434/CHEMRXIV-2023-XDV03.

(19) Szegedy, C.; Zaremba, W.; Sutskever, I.; Bruna, J.; Erhan, D.; Goodfellow, I.; Fergus, R. Intriguing Properties of Neural Networks. *2nd International Conference on Learning Representations, ICLR 2014 - Conference Track Proceedings* **2013**.

(20) Gao, P.; Han, J.; Zhang, R.; Lin, Z.; Geng, S.; Zhou, A.; Zhang, W.; Lu, P.; He, C.; Yue, X.; Li, H.; Qiao, Y. LLaMA-Adapter V2: Parameter-Efficient Visual Instruction Model. **2023**.

(21) Radford, A.; Wu, J.; Child, R.; Luan, D.; Amodei, D.; Sutskever, I. Language Models Are Unsupervised Multitask Learners. **2019**.

(22) Touvron, H.; Martin, L.; Stone, K.; Albert, P.; Almahairi, A.; Babaei, Y.; Bashlykov, N.; Batra, S.; Bhargava, P.; Bhosale, S.; Bikel, D.; Blecher, L.; Ferrer, C. C.; Chen, M.; Cucurull, G.; Esiobu, D.; Fernandes, J.; Fu, J.; Fu, W.; Fuller, B.; Gao, C.; Goswami, V.; Goyal, N.; Hartshorn, A.; Hosseini, S.; Hou, R.; Inan, H.; Kardas, M.; Kerkez, V.; Khabsa, M.; Kloumann, I.; Korenev, A.; Koura, S.; Lachaux, M.-A.; Lavril, T.; Lee, J.; Liskovich, D.; Lu, Y.; Mao, Y.; Martinet, X.; Mihaylov, T.; Mishra, P.; Molybog, I.; Nie, Y.; Poulton, A.; Reizenstein, J.; Rungta, R.; Saladi, K.; Schelten, A.; Silva, R.; Michael, E.; Ranjan, S.; Xiaoqing, S.; Tan, E.; Tang, B.; Taylor, R.; Williams, A.; Kuan, J. X.; Xu, P.; Yan, Z.; Zarov, I.; Zhang, Y.; Fan, A.; Kambadur, M.; Narang, S.; Rodriguez, A.; Stojnic, R.; Edunov, S.; Scialom, T. Llama 2: Open Foundation and Fine-Tuned Chat Models. **2023**.

(23) Gimadiev, T. R.; Lin, A.; Afonina, V. A.; Batyrshin, D.; Nugmanov, R. I.; Akhmetshin, T.; Sidorov, P.; Duybankova, N.; Verhoeven, J.; Wegner, J.; Ceulemans, H.; Gedich, A.; Madzhidov, T. I.; Varnek, A. Reaction Data Curation I: Chemical Structures and Transformations Standardization. *Mol Inform* **2021**, *40* (12), 2100119. https://doi.org/10.1002/minf.202100119.

(24) Lin, A.; Dyubankova, N.; Madzhidov, T. I.; Nugmanov, R. I.; Verhoeven, J.; Gimadiev, T. R.; Afonina, V. A.; Ibragimova, Z.; Rakhimbekova, A.; Sidorov, P.; Gedich, A.; Suleymanov, R.; Mukhametgaleev, R.; Wegner, J.; Ceulemans, H.; Varnek, A. Atom-to-atom Mapping: A Benchmarking Study of Popular Mapping Algorithms and Consensus Strategies. *Mol Inform* **2022**, *41* (4), 2100138. https://doi.org/10.1002/minf.202100138.

(25) Raffel, C.; Shazeer, N.; Roberts, A.; Lee, K.; Narang, S.; Matena, M.; Zhou, Y.; Li, W.; Liu, P. J. Exploring the Limits of Transfer Learning with a Unified Text-to-Text Transformer. *Journal of Machine Learning Research* **2020**, *21*, 1–67.

(26) Hendrycks, D.; Gimpel, K. Gaussian Error Linear Units (GELUs). **2016**. https://doi.org/10.48550/arXiv.1606.08415.

(27) Tabinda Kokab, S.; Asghar, S.; Naz, S. Transformer-Based Deep Learning Models for the Sentiment Analysis of Social Media Data. *Array* **2022**, *14*, 100157. https://doi.org/10.1016/J.ARRAY.2022.100157.